%% file: main.tex
\documentclass{article}

\PassOptionsToPackage{numbers, compress}{natbib}

\usepackage{PRIMEarxiv}
\usepackage{color}
\usepackage{natbib}
\usepackage{wrapfig}

\usepackage{amssymb}
\usepackage[utf8]{inputenc}
\usepackage[T1]{fontenc}
\usepackage{hyperref}
\usepackage{url}
\usepackage{booktabs}
\usepackage{newunicodechar}
\newunicodechar{）}{)}
\usepackage{amsfonts}
\usepackage{nicefrac}
\usepackage{microtype}
\usepackage{array}
\usepackage{multicol}
\usepackage{multirow}
\usepackage[utf8]{inputenc}
\usepackage{tabularx}
\usepackage[T1]{fontenc}
\usepackage{graphicx}
\usepackage{enumitem}
\usepackage[table,xcdraw]{xcolor}
\usepackage{amsmath}
\usepackage{fontawesome}
\usepackage{makecell}

\usepackage[utf8]{inputenc}

\definecolor{bb}{rgb}{0.12,0.565,1}
\definecolor{gg}{rgb}{0.2,0.8,0.2}
\definecolor{rr}{rgb}{1,0.85,0.2}

\newif\ifdraft
\drafttrue

\hypersetup{
    colorlinks=true,
}
\newcommand{\ours}[0]{Unified Audio Front-end LLM}
\title{UAF: A Unified Audio Front-end LLM for Full-Duplex Speech Interaction}

\author{
    Yadong Li\thanks{Core Contributors. }\enskip\enskip
    Guoxin Wu\enskip\enskip
    Haiping Hou\enskip\enskip
    Biye Li\thanks{Corresponding author.}\enskip\enskip
    \\
    Alibaba Inc.
    \\
    \texttt{\{adonlee.lyd, libiye.lby\}@alibaba-inc.com, \{guoxin.wgx,houhaiping.hhp\}@taobao.com}
    \enskip \\
}

\begin{document}

\maketitle

\input{0_abstract.tex}

\input{1_intro.tex}

\input{2_related.tex}

\input{3_method.tex}

\input{4_experiment.tex}

\input{5_conclusion.tex}

\bibliography{references.bib}
\bibliographystyle{plain}

\end{document}

%% file: 0_abstract.tex
\begin{abstract}

Full-duplex speech interaction, as the most natural and intuitive mode of human communication, is driving artificial intelligence toward more human-like conversational systems. Traditional cascaded speech processing pipelines suffer from critical limitations, including accumulated latency, information loss, and error propagation across modules. To address these issues, recent efforts focus on the end-to-end audio large language models (LLMs) like GPT-4o, which primarily unify speech understanding and generation task. However, most of these models are inherently half-duplex, and rely on a suite of separate, task-specific front-end components, such as voice activity detection (VAD) and turn-taking detection (TD). In our development of speech assistant, we observed that optimizing the speech front-end is equally crucial as advancing the back-end unified model for achieving seamless, responsive interactions. To bridge this gap, we propose the first unified audio front-end LLM (UAF) tailored for full-duplex speech systems. Our model reformulates diverse audio front-end tasks into a single auto-regressive sequence prediction problem, including VAD, TD, speaker recognition (SR), automatic speech recognition (ASR) and question answer (QA). It takes streaming fixed-duration audio chunk (e.g., 600 ms) as input, leverages a reference audio prompt to anchor the target speaker at the beginning, and regressively generates discrete tokens encoding both semantic content and system-level state controls (e.g., interruption signals). Experiments demonstrate that our model achieves leading performance across multiple audio front-end tasks and significantly enhances response latency and interruption accuracy in real-world interaction scenarios.

\end{abstract}

%% file: 1_intro.tex
\section{Introduction}
\begin{figure*}[!ht]
    \centering
    \includegraphics[width=\textwidth]{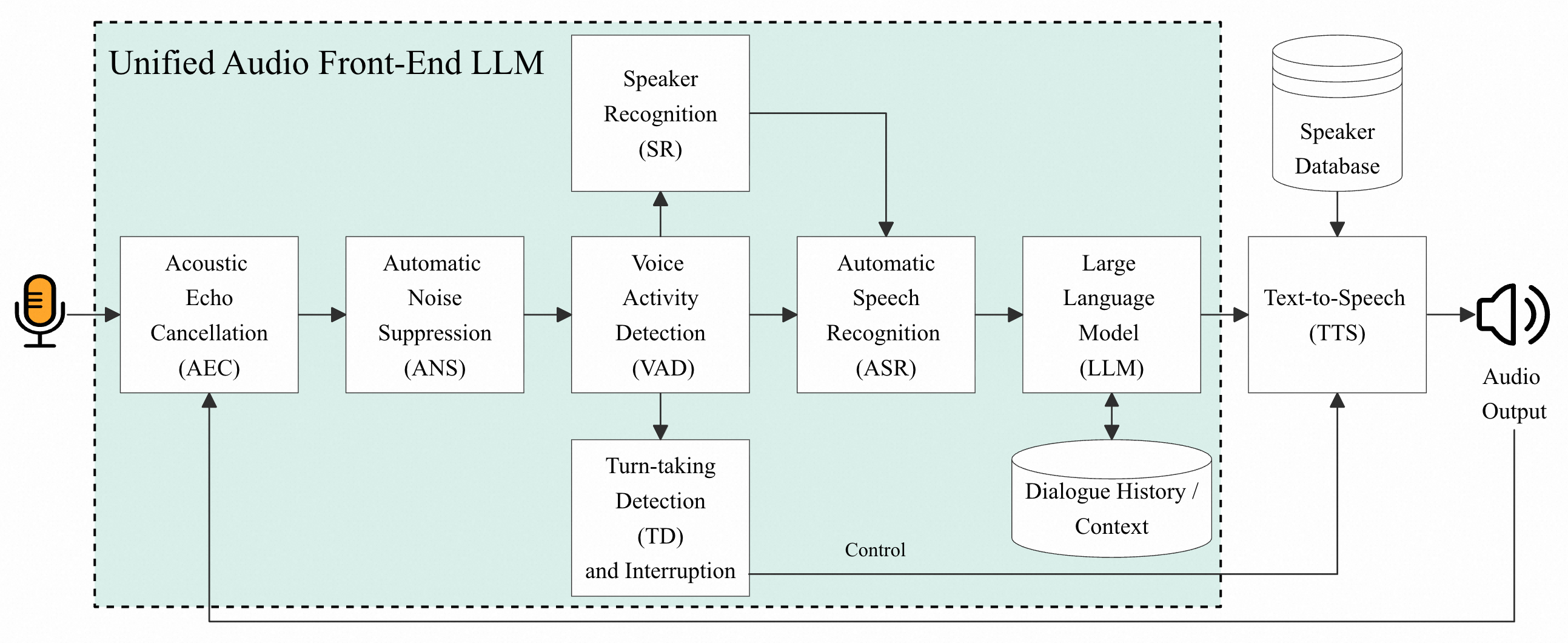}
    \caption{\textbf{\ours} and traditional cascade architecture of full-duplex systems.}
    \label{fig:architecture}
\end{figure*}

Speech interaction, as the most natural and efficient form of human communication, is driving artificial intelligence toward more human-like conversational systems. Human conversation is inherently full-duplex: participants speak, listen, and interrupt each other fluidly, relying on subtle acoustic and linguistic cues to maintain coherence and responsiveness, such as voice activity, speaker identity, turn boundaries, and contextual intent. Replicating this natural interaction in artificial systems demands not only intelligent language understanding and generation but also a robust, low-latency audio front-end capable of real-time perception under complex acoustic conditions. 

As \autoref{fig:architecture} shows, conventional speech interaction systems predominantly rely on complex cascaded pipelines, which are composed of front-end and back-end models. Raw audio first passes through a series of specialized front-end modules, including acoustic echo cancellation (AEC), automatic noise suppression (ANS), voice activity detection (VAD), speaker recognition (SR), and turn-taking detection (TD), and then will be fed into back-end models like Automatic Speech Recognition (ASR), Large Language Model (LLM) for question answer (QA), and Text-to-Speech (TTS). While it is effective to control settings, this pipeline suffers from fundamental limitations in real-world full-duplex scenarios.
\begin{itemize}[leftmargin=*]
    \item %
    First, error propagation and nonlinear distortion is unavoidable in cascaded systems, degrading overall reliability. For example, traditional signal processing front-ends tend to improve the signal-to-noise ratio (SNR) as a single goal, and noise reduction algorithms based on spectral subtraction often destroy the spectral structure of weak speech while suppressing noise, resulting in a significant decline of the ASR Model.
    \item %
    Second, the disjoint optimization of individual tasks prevents the system from leveraging cross-task dependencies and semantic-agnostic decision. For example, traditional VAD models make judgments based only on energy or spectral features, the user's tone words when thinking or non-interactive voices in the background are easily misjudged as interruption signals, resulting in frequent false interruption.
    \item %
    Third, each module in cascaded architecture introduces computational redundancy and latency accumulation. Consequently, the end-to-end delay is difficult to compress to the comfort zone of human perception (usually 200ms-500ms), making timely interruption (e.g., “stop!” during system playback) difficult to achieve. 
\end{itemize}

Recent advances in speech foundation models and end-to-end spoken language systems, exemplified by GPT-4o~\citep{HelloGPT4o}, have demonstrated remarkable progress. Many recent end-to-end models unify audio back-end tasks (i.e., speech understanding and generation task) within a single large language model (LLM) framework, where LLMs take speech representations as input and generate both text tokens and speech tokens simultaneously, such as Kimi-Audio~\citep{kimiteam2025kimiaudiotechnicalreport}, Step-Audio 2~\citep{huang2025stepaudiounifiedunderstandinggeneration}, MiMo-Audio~\citep{coreteam2025mimoaudioaudiolanguagemodels}, GLM-4-Voice~\citep{zeng2024glm4voiceintelligenthumanlikeendtoend}, Longcat-Flash-Omni~\citep{meituanlongcatteam2025longcatflashomnitechnicalreport}, Qwen2.5-Omni~\citep{xu2025qwen25omnitechnicalreport} and Qwen3-Omni~\citep{xu2025qwen3omnitechnicalreport}. These models rely on both large-scale audio-text pre-training and post-training to develop strong audio capabilities.

However, most of these models are inherently half-duplex, which rely on pluggable VAD and TD modules to support barge-in~\citep{chen2025fireredchatpluggablefullduplexvoice}, and they also usually need external front-end processing models to handle real-world challenges like far-field pickup, background noise, acoustic echo, and overlapping speech. Therefore, they still face the drawbacks of the cascading scheme mentioned earlier. In our development of speech assistant, we found that front-end robustness is as critical as back-end intelligence for user satisfaction. Delays in detecting a user’s interruption or misattribution of speaker turns can break the illusion of natural conversation, regardless of how fluent the generated response may be. This observation motivates a paradigm shift: instead of treating front-end tasks as preprocessing steps, we propose to embed them directly into a unified LLM-based generative framework.

To this end, we introduce a Unified Audio Front-End (UAF) LLM, the first large language model explicitly designed to unify core audio front-end tasks for full-duplex speech interaction. Our model reformulates VAD, TD, SR,  ASR and QA task as a single sequence-to-sequence prediction problem. It operates on streaming audio in fixed-duration chunks (e.g., 600 ms), incorporates a reference audio prompt to lock onto the target speaker, and outputs a discrete token sequence that jointly encodes semantic content (ASR text and model response) and system-level control signals (e.g., “user speaking,” “turn end,” “interruption detected”). By training on diverse real-world interaction data including simulated echo, noise, and overlapping speech, our model learns implicit cross-task representations that outperform cascaded baselines. Experiments show that our UAF achieves state-of-the-art results across individual front-end tasks. More importantly, it enables a truly integrated architecture where perception and action are co-designed within a single generative framework.

%% file: 2_related.tex
\section{Related works}
\subsection{Speech Front-end Processing in Full-duplex Systems}

As \autoref{fig:architecture} shows, traditional full-duplex speech assistants rely on a cascade of signal processing modules. Acoustic echo cancellation (AEC) is typically handled by adaptive filters or deep learning models such as DCCRN-Echo ~\citep{hu2020dccrndeepcomplexconvolution}. Voice activity detection (VAD) and speaker recognition (SR) are often implemented as separate systems. Early works on VAD relied on hand-crafted acoustic features, such as energy ratio, zero-crossing rate, and signal periodicity~\citep{zaydin2019ExaminationOE,sakhnov2009low,Junqua1991ASO}. Currently, VAD models based on neural network have gained significant attention in speech research, such as deep neural networks (DNN)~\citep{Kang2016DNNBasedVA} and feedforward sequential memory networks (FSMN)~\citep{Zhang2015FeedforwardSM,Zhang2018DeepFSMNFL}. Traditional SR frameworks typically extracts speaker embedding to represent speaker characteristics, and then employ clustering to group speaker embeddings or end-to-end neural diarization for identifying speakers~\citep{chen20243d}. Turn-taking detection (TD), though less standardized, usually leverages prosodic cues or dialogue context, such as TEN Turn Detection~\citep{tenvad}, Silero VAD~\citep{SileroVAD}, and FSMN-VAD~\citep{Zhang2018DeepFSMNFL}. However, these approaches remain confined to enhancement tasks and do not incorporate higher-level semantic or interaction-aware signals like speaker identity or turn boundaries.

The advent of large language models (LLMs) has notably advanced generative AI. Recent efforts have explored LLMs to predict state tokens for turn-taking or vad task. TurnGPT~\citep{ekstedt-skantze-2020-turngpt} injects a speaker embedding at each position and predicts the speaker ID to determine turn transitions. VITA~\citep{fu2024vita} deploys two models in paralle, one for response generation and another for continuous listening, and predicts a state token to manage turns. FlexDuo~\citep{liao2025flexduopluggableenablingfullduplex} predicts turn transitions from past context and incoming speech chunks. SpeakerLM~\citep{yin2026speakerlmendtoendversatilespeaker} introduces a unified multimodal large language model for speaker recognition (SR) and automatic speech recognition (ASR) in an end-to-end manner. Easy Turn~\citep{li2025easyturnintegratingacoustic} proposes an open-source, modular turn-taking detection model that finetune LLM backbones to predict dialogue turn states, which integrates acoustic and linguistic bimodal information.

While effective in isolation and controllable, these components are rarely co-optimized, leading to suboptimal performance under real-world conditions such as double-talk, far-field pickup, or device-induced nonlinearities. Crucially, none of these models can integrate semantic content (ASR result and model response) and system-level control signals into the same framework, leaving a gap between perception and action in spoken conversational systems.

\subsection{End-to-End Speech Large Language Models}
In recent years, end-to-end speech interaction LLMs like GPT-4o have attracted great attention for their ability to support fluent, expressive, and emotionally rich spoken interactions. Depending on whether the model can listen and speak simultaneously, which is a core characteristic of human communication, recent end-to-end speech conversational systems can be broadly categorised into two types: half-duplex (turn-based) and full-duplex speech LLMs.

Currently, most of end-to-end speech LLMs operate in a half-duplex manner, such as Kimi-Audio~\citep{kimiteam2025kimiaudiotechnicalreport}, Step-Audio 2~\citep{huang2025stepaudiounifiedunderstandinggeneration}, MiMo-Audio~\citep{coreteam2025mimoaudioaudiolanguagemodels}, GLM-4-voice~\citep{zeng2024glm4voiceintelligenthumanlikeendtoend}, Longcat-Flash-Omni~\citep{meituanlongcatteam2025longcatflashomnitechnicalreport}, Qwen2.5-Omni~\citep{xu2025qwen25omnitechnicalreport} and Qwen3-Omni~\citep{xu2025qwen3omnitechnicalreport}. While these models can engage in turn-based speech conversations, they lack an internal duplex strategy for modeling dialogue dynamics like turn-taking. Instead, they rely on external VAD modules to alternate between listening and speaking states. As a result, they struggle with key aspects of natural conversation (e.g., barge-ins and back-channel) which require the ability to listen and speak simultaneously.

Full-duplex speech LLMs can process streaming speech input and output simultaneously, and also can determine when to speak or stop. Moshi ~\citep{kyutai2024moshi} and OmniFlatten~\citep{zhang2025omniflattenendtoendgptmodel} involves injecting audio codecs into the LLM vocabulary, and they demand large-scale speech-text paired data to prevent catastrophic forgetting. The other models exampled by VITA~\citep{fu2024vita} and Freeze-Omni~\citep{wang2024freezeomnismartlowlatency} connect the LLM backbone with speech encoder and synthesizer through embeddings, without significantly hurting the LLMs. However, they are not standalone full-duplex since one LLM instance can only listen or speak, and they require two separate LLM processes to manage simultaneous listening and speaking. In contrast, SALMONN-omni~\citep{tang2024salmonn} introduces a novel duplex strategy that enables a single LLM to perform standalone full-duplex speech interaction. Currently, there is still significant room for improvement in both the limited controllability and latency of these end-to-end full-duplex models.  

Notably, none of these models treat front-end tasks as learnable components within the LLM itself. This separation forces engineers to maintain two distinct stacks: one for signal conditioning and another for audio-language reasoning. Our work bridges this gap by embedding front-end perception directly into a LLM-based generative framework.

%% file: 3_method.tex
\section{Method}
\begin{figure*}[!ht]
    \centering
    \includegraphics[width=\textwidth]{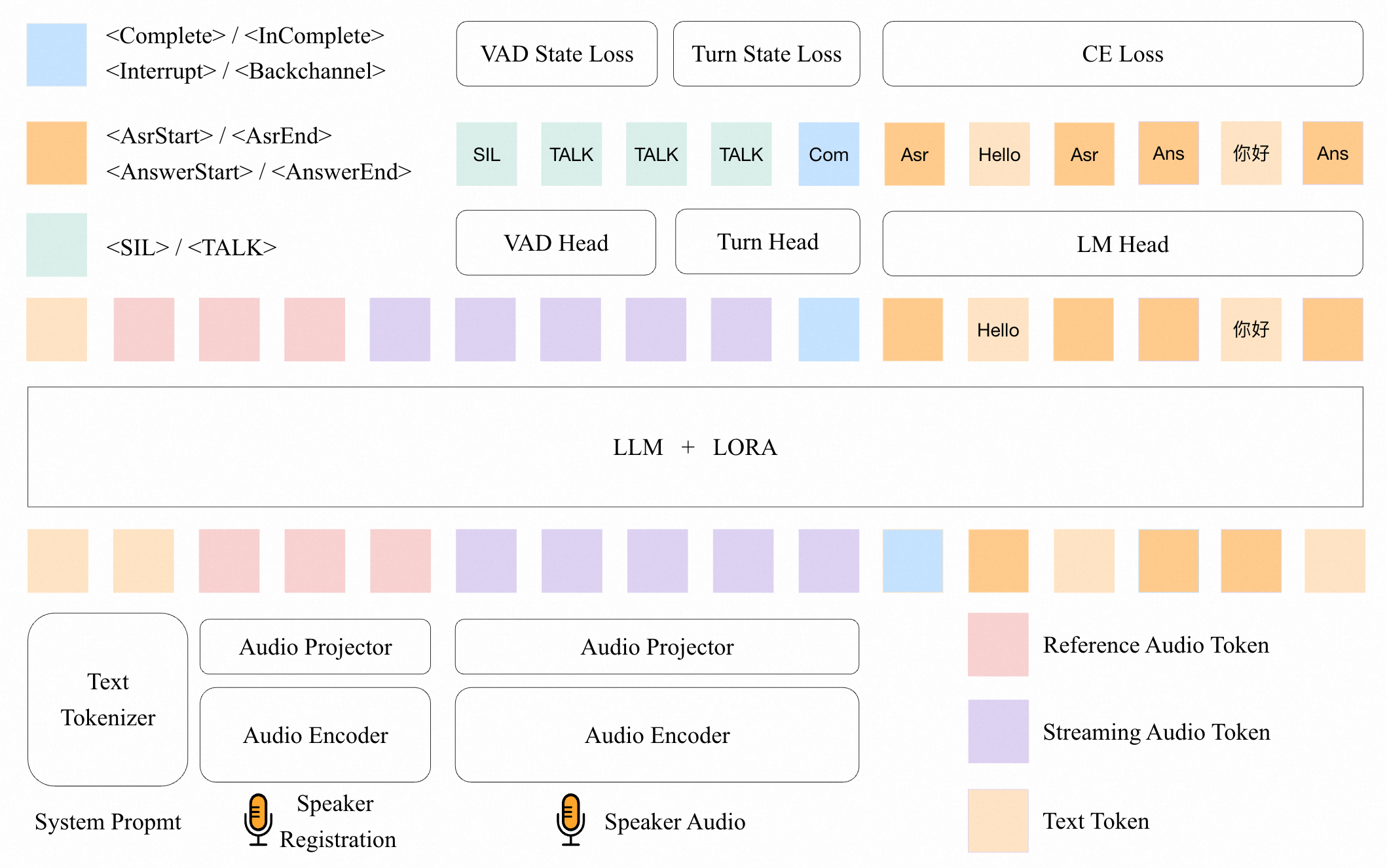}
    \caption{\textbf{Architecture of \ours.} Our model reformulates diverse front-end tasks, including speaker recognition (SR), voice activity detection (VAD), automatic speech recognition (ASR), turn-taking detection (TD), and question answer (QA), into a single sequence prediction problem. It takes streaming audio as input, processes fixed-duration segments (e.g., 600 ms) in real time, and leverages a reference audio prompt to anchor the target speaker. The output is a discrete token sequence encoding both semantic content and system-level state controls.}
    \label{fig:model-architecture}
\end{figure*}

\autoref{fig:architecture} shows the architecture of the full-duplex speech dialogue system based on the cascade scheme, in which the front-end part contains ANS, AEC, VAD, SR, ASR, TD and other tasks, and each task has a separate model. We present unified audio front-end LLM, a unified architecture that reformulates multiple audio front-end tasks as a sequence prediction problem solvable by a large language model. 

\subsection{Task Definition}
The core idea is to represent all perceptual and interaction-level information as a discrete token sequence, enabling joint modeling of semantics and system control, and the problem is formalized as follows:

\subsubsection{Problem Formulation}
Given a continuous audio chunk stream \begin{math}A_{stream}\end{math} and a reference audio \begin{math}A_{ref}\end{math} of a target speaker, our goal is to train a unified model \begin{math}\mathcal{M}\end{math} that can predict a sequence of discrete tokens at the next moment based on the current history of acoustic input.
Formally, We slice a continuous audio chunk stream into a sequence of time frames of fixed length (600ms) \begin{math}A_{stream} = \{a_1, a_2, ..., a_t\}\end{math}, and each audio chunk is followed by semantic tokens \begin{math}x_{t}\end{math} and state tokens \begin{math}s_{t}\end{math}. The training sample can be organized as:

\begin{equation}
A_{ref}, System\;Prompt, a_{1}, [x_{1};s_{1}], a_{2}, [x_{2};s_{2}], ... , a_{t}, [x_{t};s_{t}]
\end{equation}

At the moment \begin{math}t\end{math}, the joint probability distribution of the model can be defined as:
\begin{equation}
P(x_t, s_t \mid x_{\le t-1}, s_{\le t-1}, a_{\le t}, A_{ref}) = \prod_{i=1}^{L} P(x_{t,i},s_{t} \mid x_{\le t-1}, s_{\le t-1}, a_{\le t}, A_{ref})
\end{equation}
where \begin{math}x_{t,i}\end{math} indicates the \begin{math}i\end{math} token generated in the time step corresponding to the \begin{math}t\end{math} audio chunk, and \begin{math}L\end{math} indicates the maximum decoding length.

The core advantage of this unified autoregressive modeling approach lies in its implicit acoustic processing ability through attention mechanism. ANS and AEC are no longer explicit output tasks (i.e., the model does not output a denoised waveform). Instead, the model is only trained to output a control state token. For example, if the noise in the input audio is too large to match the features of target speaker, the model will predict a silence state token \begin{math}\langle SIL \rangle\end{math}. In this way, the model is forced to learn to implicitly distinguish between target signals and other interference, which can avoid the signal distortion caused by the traditional front-end process methods.

\subsubsection{Model Input and Output Space}
The input of our model consists of reference audio prompt, system prompt and streaming audio chunks, which are encoded to the hidden space of LLM through an audio encoder.
\begin{itemize}[leftmargin=*]
    \item Reference Audio Prompt:  We register the reference audio of the target speaker (usually 3-5 seconds) directly in the front of input context, which is the key to achieving speaker recognition and personalized locking. Our model encodes the reference audio to feature space, \begin{math}a_{ref} = \text{Encoder}(A_{ref})\end{math}, which serves as a Query/Key anchor in the attention mechanism, guiding the model to focus only on speech segments that match the target speaker’s voiceprint features during subsequent streaming input.
    \item Streaming Audio Chunks: The real-time streaming audio, which contains noise, reverberation, interfering vocals and echoes, serves as the context of our model. The fixed-duration streaming audio chunk (e.g., 600 ms) is encoded to streaming audio token, \begin{math}a_{t} = \text{Encoder}(A_t)\end{math}, making it possible to predict semantic content and system-level control signals in one unified model.
\end{itemize}

To unify audio front-end tasks including VAD, SR, ASR, TD, and QA in one model, we expand the vocabulary of LLM with special token of control states. The specific tasks of the model in inference are defined as follows:
\begin{itemize}[leftmargin=*]
    \item State Token. The model must predict the interactive control state of the current audio chunk \begin{math}a_{t}\end{math} at the beginning of response:
    \begin{itemize}[leftmargin=*]
        \item \texttt{[<SIL>, <TALK>]} represent the VAD states of target speaker, which can unify VAD, AEC, and target speaker recognition task. The \texttt{<SIL>} state indicates that the current audio chunk \begin{math}a_{t}\end{math} contains only background noise, echo (AEC residual), or speech from a non-target speaker. The \texttt{<TALK>} state indicates that effective speech activity of the target speaker has been detected, but a complete semantic boundary or turn-taking intent has not yet been established. 
        \item \texttt{[<Complete>, <InComplete>, <Interrupt>, <Backchannel>]} represent the TD states. The \texttt{<Complete>} state indicates that the user has fully expressed their intent and expects an immediate response from the spoken dialogue system. The \texttt{<InComplete>} state occurs when a user pauses but clearly has not finished speaking, and the full-duplex system will continue listening until the user’s semantic expression is complete, rather than interrupting prematurely. \texttt{<Backchannel>} state refers to brief listener responses (e.g., “Uh-huh.”, “Right.”) that indicate active engagement and comprehension while the speaker is talking, and it should not interrupt the system’s speech output, which is critical for maintaining interaction fluency and enhancing user experience. The \texttt{<Interrupt>} state refers to cases where users explicitly request to pause or terminate the interaction (e.g., “shut up”, “please stop”), serving as an efficient and concise way to end the system’s current turn or completely halt the dialogue.
    \end{itemize}
    \item Semantic Token. When our model detects complete semantic boundary or round of the target speaker, the model will predict ASR result based on all the previous audio chunk in the current round. The ASR result will be wrapped in special tokens \texttt{[<AsrStart>, <AsrEnd>]}, following the control status token. When the turn state is \texttt{<Complete>} or \texttt{<Interrupt>}, the model is trained to generate a response suitable for the ASR query, which is wrapped with \texttt{<AnswerStart>} and \texttt{<AnswerEnd>} special tokens.
\end{itemize}

\subsection{\ours}
\subsubsection{Model Architecture}

\autoref{fig:model-architecture} illustrates the overall architecture of \ours. Specifically, our model employs an "Encoder-Projector-LLM" architecture, which is adapted from Qwen3-Omni-30B-A3B-Instruct~\citep{xu2025qwen3omnitechnicalreport} model. It integrates audio and text generation capabilities and performs excellently on multilingual ASR and interactive tasks. For multi-speaker audio input, we use an audio encoder for encoding, followed by a projector to inject the audio embedddings into the feature space of the text LLM. It comprises the following key components:
\begin{itemize}[leftmargin=*]
\item Audio encoder converts the raw speech waveforms of the target speaker including reference audio \begin{math}A_{ref}\end{math} and each 600ms audio segment \begin{math}A_{t}\end{math} from the audio stream into high-dimensional acoustic feature representations. 
\item Audio projector maps the acoustic features output by the audio encoder to the semantic embedding space of the LLM, achieving cross-modal alignment.
\begin{equation}
a_{ref} = \text{Projector}(\text{Encoder}(A_{ref})), a_{t} = \text{Projector}(\text{Encoder}(A_{t}))
\end{equation}
\item Tokenizer: The semantic tokens \begin{math}X_t\in\mathcal{V}_{text}\end{math} of the large language model tokenizer are supplemented with state tokens \begin{math}S_t\in\mathcal{V}_{state}\end{math}, and the vocabulary is \begin{math}\mathcal{V} = \mathcal{V}_{text} \cup \mathcal{V}_{state}\end{math}.
\begin{equation}
x_{t}, s_{t} = \text{Tokenizer}(X_t, S_t)
\end{equation}
\item Large Language Model backbone is adapted from the thinker of Qwen3-Omni. LoRA (Low-Rank Adaptation)~\citep{hu2021loralowrankadaptationlarge} is used throughout for efficient fine-tuning, avoiding the computational overhead and catastrophic forgetting caused by full parameter updates.
\begin{equation}
h_t = \text{LLM\_Decoder}(x_{\le t-1}, s_{\le t-1}, a_{\le t}, a_{ref})
\end{equation}
\item LM Head, VAD Head, and Turn Head. 
\begin{equation}
x_t = \text{LM\_Head}(h_{t}), s_{t\_vad} = \text{VAD\_Head}(h_{t}), s_{t\_turn} = \text{Turn\_Head}(h_{t})
\end{equation}
Based on the audio stream and text in the historical context, the VAD head determines whether the current 600ms audio contains the target speaker's speech and generates \texttt{<SIL>} or \texttt{<TALK>}. Similarly, the turn head predicts the turn-taking state: \texttt{<Complete>, <InComplete>, <Interrupt> and <Backchannel>}. The loss function is shown in \begin{math}\mathcal{L}_{state}\end{math}.
The LM Head is the original LLM Head. Based on the audio stream and text in the historical context, it auto-regressively generates ASR results and question response. The loss function is shown in \begin{math}\mathcal{L}_{text}\end{math}.
\end{itemize}
\begin{equation}
\mathcal{L}_{text} =-\frac{1}{T} \sum_{t=1}^{T} \log P(x_{t} \mid x_{\le t-1}, a_{\le t},  a_{ref})
\end{equation}
\begin{equation}
\mathcal{L}_{state} = -\frac{1}{T} \sum_{t=1}^{T} \log P(s_{t\_vad},s_{t\_turn} \mid x_{\le t-1}, s_{\le t-1}, a_{\le t},  a_{ref})
\end{equation}
The final training loss is a weighted sum of \begin{math}\mathcal{L}_{text}\end{math} and \begin{math}\mathcal{L}_{state}\end{math}.
\begin{equation}
\mathcal{L}_{total} = \alpha\mathcal{L}_{text} + (1- \alpha)\mathcal{L}_{state}
\end{equation}

\subsubsection{Multi-stage Alignment training}
Considering the difference between learning difficulty and available data volume of various front-end tasks, we apply a multi-stage training strategy to progressively enhance the model’s audio front-end capabilities for VAD, SR, ASR, TD and QA tasks.

\begin{itemize}[leftmargin=*]
    \item \textbf{Stage I: VAD / SR / ASR Continue Pretrain.} To enhance the ASR performance of target speaker under complex conditions, we first train our UAF using 6,000 hours audio with VAD state and ASR result of target speaker, leading to a model for VAD, SR and ASR tasks. To efficiently adapt the LLM while preserving its pre-trained language capabilities, we apply the LoRA strategy with a learning rate of $1e-4$, and the newly added VAD head which is initialized from the LM head also participates in the training. 
    \item \textbf{Stage II: TD and QA Alignment.} We use 1,000 hours audio for TD and QA task to enhance its turn-taking detection ability, while 1,000 hours audio sampled from Stage I's data is also used to preserve its original capabilities. When the turn state is \texttt{<Complete>} or \texttt{<Interrupt>}, the model is trained to generate a response suitable for the ASR query, which is wrapped with \texttt{<AnswerStart>} and \texttt{<AnswerEnd>} tags. We still keep both the LLM and audio encoder frozen in this stage, and train the newly added Turn Head and LoRA parameters.
    \item \textbf{Stage III: All-Task Training.} Finally, we use multi-turn user–agent dialogues data including all tasks to jointly fine-tune previous trainable modules while applying LoRA to the LLM. This stage enables jointly optimization to better integrate linguistic and acoustic information in complex practical acoustic conditions.
\end{itemize}

\section{Full-duplex Interaction Data Synthesis Pipeline}
\label{data}

\begin{figure*}[!ht]
    \centering
    \includegraphics[width=\textwidth]{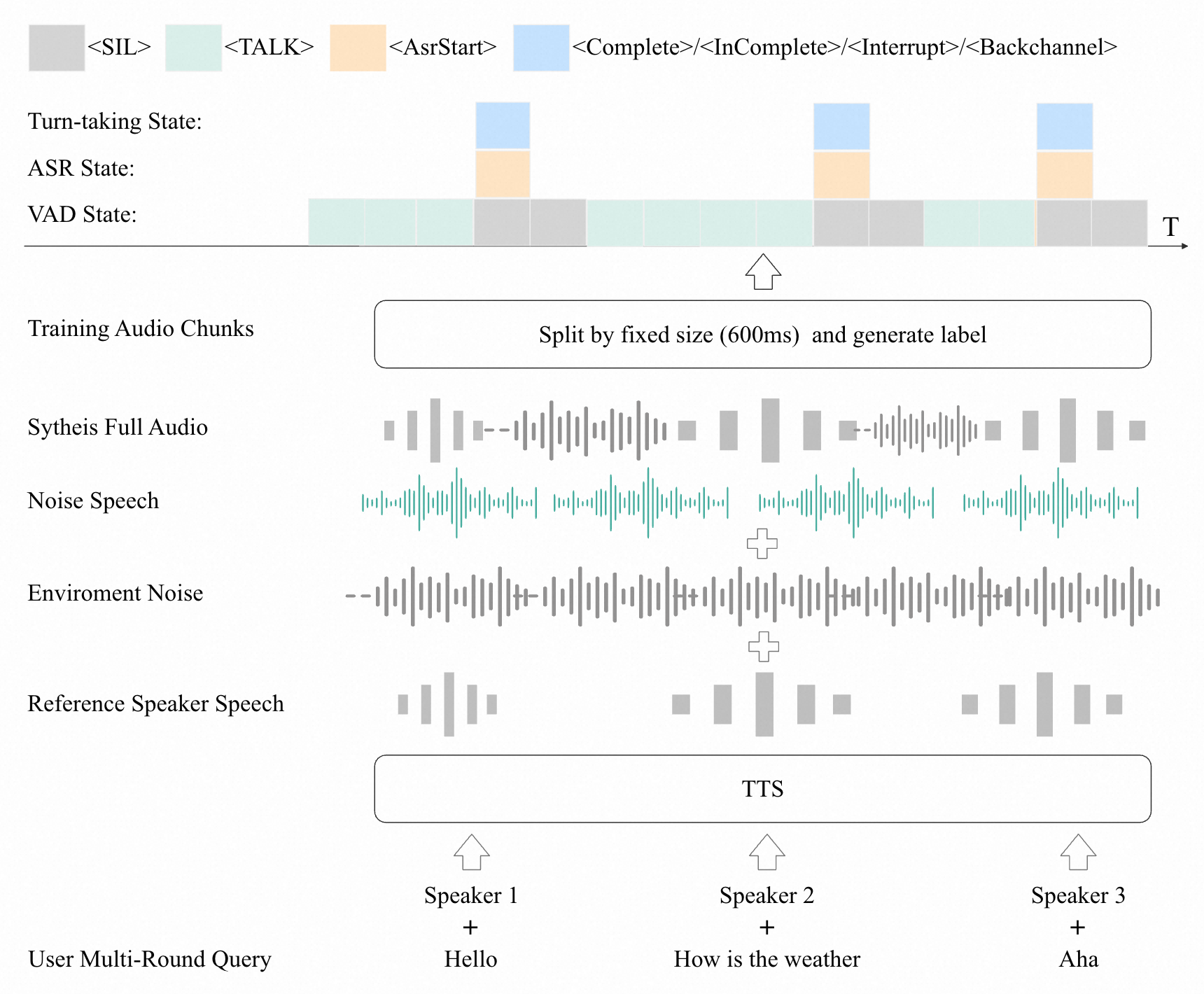}
    \caption{\textbf{Full-duplex Interaction Data Synthesis Pipeline of \ours.}}
    \label{fig:data_pipeline}
\end{figure*}

Real-world full-duplex data of human–agent interaction is extremely scarce. To train a robust unified audio front-end model for full-duplex speech interaction, we construct a hybrid data pipeline that combines real-world recordings and large-scale synthetic dialogues. This pipeline yields realistic, multi-talker far-field data suitable for challenging conditions. 

\subsection{Data Source Composition}
The input of our model is a mixed audio signal containing clean speech of target speaker, interference speech and environmental acoustics, while the ground-truth label is derived exclusively from the target speaker. This can force the model to learn to suppress irrelevant acoustic content (e.g., echo, background talkers) and attend only to the target speaker, especially when guided by a reference audio prompt. 

\begin{itemize}[leftmargin=*]
    \item Clean Speech. We curate clean speech from both public datasets and in-house collections, which serves as the source of target speaker and reference prompt. We collect large-scale mandarin speech corpora for linguistic diversity from public datasets, including Fleurs~\citep{fleurs2022arxiv}, AISHELL-1~\citep{bu2017aishell1opensourcemandarinspeech}, AISHELL-2~\citep{du2018aishell2transformingmandarinasr}, KeSpeech~\citep{tang2021kespeech}, and WenetSpeech~\citep{zhang2022wenetspeech10000hoursmultidomain}. We also extracts over 1,000 hours in-house multi-speaker audio from publicly available podcasts. 
    \item Interference speech data refers to the competing voices in cocktail-party scenarios, which is synthesized from VoxCeleb~\citep{Nagrani_2017} and CommonVoice~\citep{ardila2020commonvoicemassivelymultilingualspeech} datasets.
    \item Environmental acoustic data (such as background noise and reverberation) is sampled from MUSAN~\citep{snyder2015musanmusicspeechnoise} dataset.
\end{itemize}
\subsection{Synthesis Pipeline}
\autoref{fig:data_pipeline} shows our data pipeline for synthetic dialogues.
\begin{itemize}[leftmargin=*]
    \item Dialogue Generation: We use real-world recordings and large-scale synthetic dialogues to generate dialogue-like speech sequences, reflecting real speech assistant scenarios.
    \begin{itemize}[leftmargin=*]
        \item Real data. From near-field clean utterances in public datasets, we randomly sample 2–4 speakers and concatenate their utterances into ~50-second dialogue-like sequences.
        \item Synthetic data. First, we instruct a large language model to generate multi-turn user–agent dialogues. Second, a target speaker identity is selected from a pre-enrolled voice bank. Using a zero-shot voice cloning TTS system (CosyVoice~\citep{du2025cosyvoice}), each user turn is synthesized with consistent (or deliberately inconsistent) voice characteristics across turns, enabling explicit training of speaker consistency modeling.
    \end{itemize}
    \item Realistic Interaction Simulation: The synthesized user turns are assembled into continuous audio streams with:
    \begin{itemize}[leftmargin=*]
        \item Natural pauses: Silent gaps (0.5–3 s) inserted before/after each user utterance to mimic real user behavior (e.g., listening to system response or browsing).
        \item Environmental noise: Additive noise from MUSAN~\citep{snyder2015musanmusicspeechnoise} at varying SNRs (0–20 dB).
        \item Competing talkers: Non-target speaker utterances overlaid at random positions to challenge speaker discrimination.
        \item Echo injection: To emulate system playback, we synthesize agent responses using TTS and convolve them with measured or simulated electro-acoustic transfer functions (including nonlinear distortions). These echo signals are then mixed into the input during user speaking segments, particularly in barge-in scenarios.
    \end{itemize}
\end{itemize}

\subsection{Streaming Sample Construction}
The final continuous audio obtained by our dialogue-like synthesis pipeline is segmented into fixed-duration chunks (e.g., 600 ms) to match the model’s streaming inference window. For each chunk, we generate a ground-truth token sequence based on its acoustic and semantic content. \autoref{tab:training_scenarios} summarizes key training scenarios. This structured labeling strategy enables the model to jointly learn semantic transcription and interaction-aware control signals within a single autoregressive framework.

To obtain high-precision word-level timestamps for training label alignment and token sequence generation, we propose an acoustic-aware timestamp extraction pipeline built upon the speech recognition model Paraformer-Zh~\citep{gao2022paraformer}. The procedure consists of four stages: 
\begin{itemize}[leftmargin=*]
    \item Coarse Timestamp Extraction: We first apply the Paraformer-Zh model to predict initial word-level time boundaries (start and end times) as a coarse estimate. While efficient, these timestamps often suffer from misalignment due to model latency or ambiguous phonetic transitions.
    \item Audio Slicing: Using the coarse timestamps, we segment the original waveform into short audio clips which represent a single recognized character.
    \item Acoustic Analysis: For each audio clip, we compute two low-level acoustic features: short-time energy (as a proxy for loudness) and zero-crossing rate (indicative of high-frequency content and voicing). A dynamic threshold, adaptively estimated from the global energy distribution of the utterance, is used to distinguish voiced segments from genuine "voices" or "silence" within each clip.
    \item Boundary Refinement: We locate the precise start and end points of vocal activity within each segment, and then apply a small symmetric padding (±10–20 ms) to the start and end points to ensure natural auditory continuity and mitigate truncation artifacts.
\end{itemize}
    
For VAD task, we construct about 7,000 hours audio training samples with VAD state and ASR result. Based on the timestamps of the target speaker's clean audio and the inserted interaction signals, we generate corresponding VAD state \texttt{[<SIL>, <TALK>]} for each audio chunk depending on whether the current chunk contains the target speaker's voice. By the way, our timestamp extraction method significantly improves temporal accuracy. On our internal evaluation set, it achieves 4.8 times higher precision compared to WhisperX\citep{bain2022whisperx}, enabling more reliable alignment between audio chunks and discrete VAD state tokens during training. 

For ASR task of target speaker, it does not require generating labels for each audio chunk; instead, we insert the corresponding ASR result of target speaker at the audio chunk position where the transition from \texttt{<TALK>} to \texttt{<SIL>} occurs. Similarly, TD task inserts the corresponding turn state before the ASR result. Since no reliable open-source speech annotation tools exist for TD task, we design prompts to label turn states \texttt{[<Complete>, <InComplete>, <Interrupt>, <Backchannel>]} with Qwen3 LLM. Finally, we construct about 1,000 hours of audio data with turn states based on the previous VAD and ASR dataset. 

For QA task, we construct over 50k training samples. When the turn state is \texttt{<Complete>} or \texttt{<Interrupt>}, we invoke the Qwen3 LLM to generate a response suitable for the ASR query. This response is then wrapped with \texttt{<AnswerStart>} and \texttt{<AnswerEnd>} tags and appended to the ASR result, thereby enabling the model with conversational capabilities.

\begin{table}[t]
\centering
\caption{Training scenarios and corresponding ground-truth token sequences in the synthetic data pipeline.}
\label{tab:training_scenarios}
\begin{tabular}{p{3.4cm}p{4.4cm}p{7.2cm}}
\toprule
\textbf{Scenario} & \textbf{Physical Signal Composition} & \textbf{Target Token Sequence (Ground Truth)} \\
\midrule
Pure silence or noise & Noise only & \texttt{[<SIL>]} \\
\addlinespace
Interference speaker & Non-target speaker + noise & \texttt{[<SIL>]} \\
\addlinespace
Normal interaction & Target speaker + noise & \texttt{[<TALK>, <AsrStart>query<AsrEnd>, <Complete><AnswerStart>answer<AnswerEnd> or <InComplete>]} \\
\addlinespace
Intelligent barge-in & Target speaker + system echo + noise (overlap) & \texttt{[<TALK>, <AsrStart>query<AsrEnd>, <Interrupt><AnswerStart>answer<AnswerEnd> or <Backchannel>]} \\
\bottomrule
\end{tabular}
\end{table}

%% file: 4_experiment.tex
\section{Experiment}

\subsection{Major Performance} 

We conduct comprehensive experiments to evaluate the proposed Unified Audio Front-end LLM (UAF) on core front-end tasks, including voice activity detection (VAD), standard automatic speech recognition (ASR), speaker-aware ASR, and turn-taking detection (TD) with noisy multi-talker conditions. All models are evaluated under identical streaming settings (600 ms chunks). The evaluation results collectively demonstrate that unifying front-end tasks into a single LLM can preserve individual task performance, especially in complex, real-world interaction scenarios.

\subsubsection{Voice Activity Detection (VAD)}\label{sec: vad-performance}

We construct a challenging VAD evaluation set based on the WenetSpeech corpus, which includes diverse acoustic conditions such as background music, overlapping speech, and device-induced artifacts. We compare UAF against three open-source baselines: TEN-VAD~\citep{tenvad}, Silero-VAD~\citep{SileroVAD}, and FSMN-VAD~\citep{Zhang2018DeepFSMNFL}.

As shown in \autoref{tab:vad}, UAF achieves the highest F1-score (97.57\%) and recall (97.99\%) compared with three open-source VAD models, indicating its superior sensitivity to true speech segments, which is critical for reliable interruption detection in full-duplex systems. This high recall comes without severe precision degradation, demonstrating effective noise suppression via reference-prompt conditioning. The identical metrics confirm that unifying front-end tasks does not compromise VAD performance.

\begin{table}[ht]
\centering
\caption{VAD performance comparison on internal test set.}
\label{tab:vad}
\begin{tabular}{lcccc}
\toprule
Model & Accuracy(\%) & Precision(\%) & Recall(\%) & F1-score(\%) \\
\midrule
FSMN-VAD & 91.13 & 91.07 & 97.79 & 94.31 \\
Silero-VAD & 95.56 & 98.35 & 96.62 & 97.48 \\
TEN-VAD & 94.79 & 96.32 & 97.87 & 97.09 \\
UAF-30B-A3B (Ours) & 95.67 & 97.16 & \textbf{97.99} & \textbf{97.57} \\
\bottomrule
\end{tabular}
\end{table}

\subsubsection{Standard ASR Performance}\label{sec: standard-asr-performance}
We evaluate standard ASR performance on three public mandarin datasets under standard conditions (no reference audio prompt provided):
Fleurs-\textit{zh}, AISHELL-1, AISHELL-2, which cover diverse domains with varying accents and noise levels. In addition, we construct Online-test dataset based on real mobile recordings from the Taobao APP. We use Word Error Rate (WER) as evaluation metrics, and compare UAF against five open-source baselines: Paraformer-zh-streaming~\citep{gao2022paraformer}, Qwen3-Omni-30B-A3B~\citep{xu2025qwen3omnitechnicalreport}, Qwen2.5-Omni-7B~\citep{xu2025qwen25omnitechnicalreport}, Kimi-Audio-7B~\citep{kimiteam2025kimiaudiotechnicalreport}, and Qwen2-Audio-7B~\citep{chu2024qwen2audiotechnicalreport}.

Results in \autoref{tab:asr_clean} show that UAF achieves competitive WER result. Notably, UAF attains 2.43 WER on AISHELL-2, outperforming recent multimodal LLMs such as Kimi-Audio. On the challenging Online-test set, UAF reduces WER to 13.75, surpassing Qwen3-Omni-30B-A3B, demonstrating strong robustness even without speaker anchoring.

\begin{table}[ht]
\centering
\caption{Standard ASR performance (no reference audio).}
\label{tab:asr_clean}
\begin{tabular}{lcccccc}
\toprule
Model / WER & AISHELL-1 & AISHELL-2 & Fleurs-zh & Online-test \\
\midrule
Paraformer-zh-streaming & 3.05 & 3.77 & 5.98 & 23.60 \\
Qwen3-Omni-30B-A3B & 1.03 & 2.47 & 2.88 & 17.83 \\
Qwen2.5-Omni-7B & 1.13 & 2.56 & 2.92 & 19.39 \\
Kimi-Audio-7B & \textbf{0.61} & 2.56 & \textbf{2.87} & 21.93 \\
Qwen2-Audio-7B & 1.52 & 3.08 & 3.63 & 22.56 \\
UAF-30B-A3B (Ours) & 0.84 & \textbf{2.43} & 2.92 & \textbf{13.75} \\
\bottomrule
\end{tabular}
\end{table}

\subsubsection{Speaker-Aware ASR Performance with Reference Audio}\label{sec: speaker-aware-asr-performance}
To validate UAF’s ability to leverage reference audio for target-speaker focusing, we construct six speaker-conditioned ASR benchmarks by augmenting clean utterances from AISHELL-1 and AISHELL-2 with interfering speakers from VoxCeleb~\citep{Nagrani_2017} and environmental noise from MUSAN~\citep{snyder2015musanmusicspeechnoise} at varying SNRs (0-20 dB). For open-source audio LLMs with instruction-following capabilities, including Qwen3-Omni-30B-A3B~\citep{xu2025qwen3omnitechnicalreport}, Qwen2.5-Omni-7B~\citep{xu2025qwen25omnitechnicalreport}, and Kimi-Audio-7B~\citep{kimiteam2025kimiaudiotechnicalreport}, we provide a reference audio (5-second enrollment of the target speaker) within the prompt, instructing the model to identify and extract the speech content of that target speaker based on the provided reference audio.

As shown in \autoref{tab:asr_speaker_aware}, UAF dramatically outperforms all baselines across all SNR levels, indicating that existing audio LLMs still suffer from significant shortcomings in terms of speaker recognition. Even at extreme noise (2 dB), UAF achieves 5.34 WER, while Qwen3-Omni-30B-A3B suffers from 38.6 WER, a 7× relative improvement. On the augmented test set with random noise (0-10 dB), UAF achieves 3.09 WER, far surpassing Qwen3-Omni-30B-A3B (68.01) and Kimi-Audio (62.7). This confirms that UAF effectively suppresses non-target speakers and system echo when guided by a reference prompt, enabling reliable operation in realistic full-duplex scenarios.

\begin{table}[ht]
\centering
\caption{Speaker-aware ASR performance under varying SNR conditions with reference prompt.}
\label{tab:asr_speaker_aware}
\begin{tabular}{lccccc c}
\toprule
Model / WER & 2 dB & 5 dB & 10 dB & 15 dB & 20 dB & Random (0-10 dB) \\
\midrule
Qwen3-Omni-30B-A3B & 38.60 & 21.95 & 6.24 & 2.16 & 2.01 & 68.01 \\
Qwen2.5-Omni-7B & 81.77 & 70.91 & 66.66 & 67.79 & 71.00 & 102.69 \\
Kimi-Audio-7B & 36.25 & 15.35 & 4.70 & 2.07 & 1.43 & 62.70 \\
UAF-30B-A3B (Ours) & \textbf{5.34} & \textbf{2.27} & \textbf{1.43} & \textbf{1.30} & \textbf{1.24} & \textbf{3.09} \\
\bottomrule
\end{tabular}
\end{table}

\subsubsection{Turn-taking Detection Performance}\label{sec: turn-taking-performance}
To evaluate the model’s ability to understand conversational dynamics, we conduct a turn-taking detection experiment on the TD test set of Easy-Turn~\citep{li2025easyturnintegratingacoustic}, containing four types of user behaviors: \texttt{<Complete>, <InComplete>, <Interrupt>, <Backchannel>}. We compare UAF against two open-source baselines: Smart Turn V2, and Easy-Turn model.

As shown in \autoref{tab:turn_taking}, our proposed UAF model achieves state-of-the-art performance across all categories, demonstrating exceptional sensitivity to both explicit and implicit turn signals. Notably, UAF attains 100.0\% accuracy on \texttt{<Interrupt>} type, crucial for responsive full-duplex interaction, and 95.7\% on \texttt{<BackChannel>} type, significantly outperforming Qwen3-Omni-30B-A3B (28.0\%) and the Smart Turn V2 baseline (which does not support backChannel or interrupt). Even on fine-grained distinctions like \texttt{<InComplete>} type (where users trail off or hesitate), UAF achieves 98.95\% accuracy, surpassing Easy Turn (97.67\%) and Qwen3-Omni-30B-A3B (92.33\%). This indicates that the model effectively leverages acoustic cues (e.g., energy drop, pause duration) and semantic context jointly encoded in its token stream. The strong performance on \texttt{<Complete>} type (96.48\%) further confirms that UAF maintains high precision in standard scenarios while excelling in challenging, interaction-critical cases. These results validate that unifying turn-taking detection with other front-end tasks within an LLM framework enables richer modeling of conversational pragmatics.

\begin{table}[ht]
\centering
\caption{TD Performance on Easy-Turn test set.}
\label{tab:turn_taking}
\begin{tabular}{lccccc c}
\toprule
Model / Accuracy & Complete(\%) & InComplete(\%) & Backchannel(\%) & Interrupt(\%) \\
\midrule
Smart Turn V2 & 78.67 & 62.00 & - & - \\
Easy Turn & 96.33 & 97.67 & 91.00 & 98.00 \\
Qwen3-Omni-30B-A3B & 91.33 & 92.33 & 28.00 & 18.00 \\
UAF-30B-A3B (Ours) & \textbf{96.48} & \textbf{98.95} & \textbf{95.70} & \textbf{100.00} \\
\bottomrule
\end{tabular}
\end{table}

\subsection{Ablation Study}\label{sec: abla_study}
To better understand the design choices behind UAF, we conduct ablation studies on three critical aspects: model size, fine-tuning strategy, and front-end task head architecture.

\subsubsection{Model Size}\label{sec: model-scale}
We train variants of UAF based on three backbone sizes from the Qwen-Omni series: 3B, 7B, and 30B-A3B. All models are evaluated on the speaker-aware ASR benchmark with reference audio prompts under varying SNR conditions (2–20 dB). As shown in \autoref{tab:scale_ablation}, model capacity has a profound impact on robustness in low-SNR regimes, thus 3B model performs the worst. While 7B and 30B-A3B model achieve comparable WER at high SNR (e.g., 1.24 of 7B model and 1.26 of 30B-A3B model at 20 dB), the performance gap widens dramatically as noise increases. At 2 dB SNR, the 30B-A3B model achieves 5.34 WER, significantly outperforming the 7B (15.03) and 3B (38.24) variants, demonstrating that larger models better leverage reference audio prompts to suppress interference and recover target speaker's speech. Similar trends hold on the test set with random noise (0-10 dB), where 30B-A3B model attains 3.09 WER versus 15.21 WER for 3B model. These results justify our choice of the 30B-A3B backbone for deployment in our full-duplex systems.

\begin{table}[ht]
\centering
\caption{Speaker-aware ASR WER of model size ablation under varying SNR.}
\label{tab:scale_ablation}
\begin{tabular}{lcccccc}
\toprule
Model / WER & 2 dB & 5 dB & 10 dB & 15 dB & 20 dB & Random (0-10 dB) \\
\midrule
UAF-3B & 38.24 & 14.43 & 5.11 & 3.38 & 2.90 & 15.21 \\
UAF-7B & 15.03 & 5.15 & 1.92 & 1.54 & 1.26 & 5.96 \\
UAF-30B-A3B & 5.34 & 2.27 & 1.43 & 1.30 & 1.24 & 3.09 \\
\bottomrule
\end{tabular}
\end{table}

\subsubsection{Full Fine-tuning vs. LoRA}\label{sec: lora-ablation}
We compare full parameter fine-tuning with LoRA fine-tuning to balance performance, training cost, and preservation of the base model’s general capabilities (e.g., instruction following). Results in \autoref{tab:lora_ablation} show that LoRA fine-tuning achieves nearly identical performance to full parameter fine-tuning across both standard ASR and speaker-aware ASR benchmarks. On standard ASR sets (AISHELL-1/2), the WER differences between two fine-tuning methods are within 0.1. In minor noisy conditions, LoRA incurs only a minor degradation (e.g., 0.08 WER at 15 dB). Given its drastically reduced memory footprint and training time, and it avoids catastrophic forgetting of pre-trained knowledge, we adopt LoRA for all final experiments.

\begin{table}[ht]
\centering
\caption{Comparison of full fine-tuning and LoRA. Top: standard ASR; Bottom: speaker-aware ASR.}
\label{tab:lora_ablation}
\begin{tabular}{lcccccc}
\toprule
Model / WER & AISHELL-1 & AISHELL-2 & Fleurs-zh & Online-test & - & - \\
\midrule
Full FT & 0.80 & 2.40 & 2.89 & 12.90 & - & - \\
LoRA & 0.84 & 2.43 & 2.92 & 13.75 & - & - \\
\midrule
Model / WER & 2 dB & 5 dB & 10 dB & 15 dB & 20 dB & Random (0-10 dB) \\
\midrule
Full FT & 5.94 & 2.31 & 1.54 & 1.22 & 1.17 & 2.98 \\
LoRA & 5.34 & 2.27 & 1.43 & 1.30 & 1.24 & 3.09 \\
\bottomrule
\end{tabular}
\end{table}

\subsubsection{Shared LM Head vs. Dedicated Task Heads}\label{sec: unified-head-ablation}
In our interaction protocol, with the audio stream input, we expect the model generate the ASR result only when the model detects that the user has spoken a relatively complete semantic sentence, and only generate the VAD state before that. A key design question is whether VAD and TD should share the main language modeling (LM) head (i.e. shared LM head) or use dedicated lightweight heads (i.e. dedicated task heads).

In the setting of shared LM head, the model generates VAD/TD states and ASR tokens from the same decoder head. Our experiments show that this way leads to undesired coupling: every audio chunk triggers both VAD state and partial ASR output, resembling conventional streaming ASR. Consequently, the model cannot wait for a complete user utterance before committing to an ASR hypothesis, violating our interaction protocol. For TD task, this setup biases predictions toward \texttt{<Complete>} type (due to semantic emphasis), severely degrading detection precision of \texttt{<BackChannel>} and \texttt{<Interrupt>} type.

Therefore, we apply dedicated task heads in our UAF. We add two linear heads (one for VAD, one for TD) that operate independently of the LM head, which are initialized from the original LM head. The VAD head continuously monitors speaker activity, and only when a talk-to-silence (i.e. \texttt{<TALK>} to \texttt{<SIL>}) transition is detected, our full-duplex speech model triggers ASR decoding over the cached context, and the TD head will explicitly predicts four turn types. As shown in Section 5.1.4, this decoupled design enables state-of-the-art turn detection while maintaining low-latency ASR. It also aligns with human-like listening behavior: perceive first, transcribe later.

%% file: 5_conclusion.tex
\section{Conclusion}

In this work, we challenge the long-standing paradigm of modular, cascaded front-end processing in full-duplex speech systems and propose UAF (Unified Audio Front-end LLM), the first large language model that unifies core audio front-end tasks into an end-to-end generative framework. By reformulating voice activity detection (VAD), speaker recognition (SR), automatic speech recognition (ASR), turn-taking detection (TD), and question answer (QA) as a sequence prediction problem over discrete tokens, UAF enables joint modeling of semantic content and interaction-level control signals. Crucially, it leverages a reference audio prompt to anchor the target speaker, allowing robust operation in noisy multi-talker environments with system playback. Extensive experiments demonstrate that UAF not only achieves state-of-the-art performance on individual front-end tasks but also significantly enhances real-world interaction quality. It matches or exceeds existing VAD models, TD models, and leading ASR models on standard benchmarks, and notably achieves dramatic gains in speaker-aware scenarios under low SNR conditions. Our work bridges the gap between signal-level perception and language-level reasoning, paving the way for truly integrated conversational agents where “listening” is no longer a preprocessing step, but an intelligent, context-aware capability embedded within the language model itself. We hope this paradigm inspires future research toward unified perception-generation architectures for embodied and interactive AI.